\documentclass{IEEEtran}

\usepackage{amssymb, amsmath, verbatim, amsthm,url, multirow,fullpage,mathtools, graphicx, subcaption, csvsimple, outlines}
\usepackage{booktabs}

\usepackage{soul}
\usepackage[colorinlistoftodos,textsize=footnotesize]{todonotes}

\setstcolor{red}
\newcommand{\cI}{\mathcal{I}}
\newcommand{\cR}{\mathcal{R}}
\newcommand{\cL}{\mathcal{L}}

\newcommand{\R}{\mathbb{R}}
\newcommand{\C}{\mathbb{C}}

\def\ba #1\ea{\begin{align} #1 \end{align}}
\def\bas #1\eas{\begin{align*} #1 \end{align*}}
\def\bml #1\eml{\begin{multline} #1 \end{multline}}
\def\bmls #1\emls{\begin{multline*} #1 \end{multline*}}
\newtheorem{thm}{Theorem}[section]

\newtheorem{cor}[thm]{Corollary}

\theoremstyle{remark}

\theoremstyle{definition}
\newtheorem{dfn}[thm]{Definition}

\usepackage{authblk}

\title{Eigenvalues of Autoencoders in Training and at Initialization}

\author[1]{Benjamin Dees}
\author[2]{Susama Agarwala}
\author[2]{Corey Lowman}

\affil[1]{Johns Hopkins University, Mathematics Department}
\affil[2]{Johns Hopkins University Applied Physics Lab}

\date{\today}                     %% if you don't need date to appear
\begin{document}

\maketitle

\begin{abstract}
    In this paper, we investigate the evolution of autoencoders near their initialization.  In particular, we study the distribution of the eigenvalues of the Jacobian matrices of autoencoders early in the training process, training on the MNIST data set. We find that autoencoders that have not been trained have eigenvalue distributions that are qualitatively different from those which have been trained for a long time ($>$100 epochs).  Additionally, we find that even at early epochs, these eigenvalue distributions rapidly become qualitatively similar to those of the fully trained autoencoders.  We also compare the eigenvalues at initialization to pertinent theoretical work on the eigenvalues of random matrices and the products of such matrices.
\end{abstract}

\section{Introduction}

While there is a large body of literature studying the geometric properties of input space (for instance, see \cite{ID, moreID, UMAP, tSNE, Peterson}), there is less work in understanding the geometric characterizations of that a neural network has learned \cite{gag, xu2021how}, and to our knowledge, none comparing the geometric properties of a neural network at early stages of training to the geometry of the fully trained network. It is accepted in the machine learning community that the early portion of a neural network's training is an important period of change (see, for instance, \cite{FrankleEtAl}). Therefore, we wish to study, geometrically, how this change evolves over the training period. We hope that this line of inquiry will lead to insight into the training process, separating the stable geometry from the underlying stochasticity of learning. 

The authors of \cite{gag}, study autoencoders as functions from their input spaces to the reconstruction spaces.  To understand the geometry of the feature map learnt by the autoencoder, they considered the matrix of derivatives of the autoencoder, or the Jacobian matrix. For any $x$ in the input space, they consider the eigenvalues of the Jacobian matrix at $x$, which reveal geometric properties of how the network is transforming the input geometry to a feature representation of the data. While that work focused on fully trained autoencoders, in contrast, in this paper focuses on the geometry of autoencoders near initialization and how this changes with training. In the end, we wish to understand what geometric features of the autoencoder were present initially, and when various geometric aspects of the trained autoencoder emerge in the training process.

At initialization, an autoencoder's Jacobian matrices are determined randomly by the entries of the weight matrices, the bias vectors, and the particular choices of nonlinearities in the autoencoder's architecture.  Therefore, to make predictions about the distribution of eigenvalues at initialization, we draw on theoretical results pertaining to the eigenvalues of random matrices, or to products of random matrices.  Then, we compare these predictions to our empirical observations at epoch $0$.  We also observe the evolution of the distribution of eigenvalues over the early portion of the training process.

We find that the theoretical predictions about the eigenvalues are generally larger in absolute value than the empirical observations at epoch $0$, but do qualitatively describe some features of the distributions.  Additionally, we observe considerable changes in these distributions early in the training process.  In particular, while the authors in \cite{gag} observe that the trained autoencoders do have eigenvalue distributions reminiscent of projection matrices, in this paper we see that this is a result of training, not present at initialization.

\section{Theory}
An autoencoder is a neural network consisting of two components.  The {\em encoder} maps from a high-dimensional input space, $\cI = \R^N$, to a lower-dimensional latent space, $\cL = \R^d$ (where $d\ll N$).  The {\em decoder} maps the latent space to the reconstruction space $\cR = \R^N$, which is equivalent to the input space.  We denote the encoder by $f_{enc}$, and the decoder by $f_{dec}$.  The autoencoder is constructed so that the image of $f_{dec}\circ f_{enc}:\cI \to \cR$ is at most $d$ dimensional.

For an input point $x\in\cI$, we denote the corresponding reconstructed point by $y:=f_{dec}\circ f_{enc} (x)\in\cR$ and the latent representation by $z:=f_{enc}(x)\in\cL$.  The Jacobian matrix (or derivative matrix) of the autoencoder, $J_\cI(x) := D (f_{dec} \circ f_{enc})(x) = D f_{dec}(z) \cdot D f_{enc}(x)$, is a linear map from the tangent space of $\cI$ at $x$, to the tangent space of the reconstruction space at $y$.  Because the autoencoder's image is at most $d$ dimensional, for any point $x$, the Jacobian matrix $J_\cI(x)$ has at most $d$ nonzero eigenvalues.  These eigenvalues describe how much the autoencoder distorts its corresponding eigenspace---an eigenvalue of $0$ represents an eigenspace that is mapped entirely to one point, an eigenvalue of $1$ represents an eigenspace that is left unchanged.

Some eigenvalues may be complex, but because our matrices are real-valued, these complex eigenvalues come in complex conjugate pairs.  We usually consider these in polar coordinates $\lambda=re^{i\theta}$; the absolute value or ``modulus" is the $r$ here and the ``argument" is the angle $\theta$.  In this paper, when we graph distributions of the arguments, we usually graph $|\theta|$; this is because complex conjugate pairs come in the form $re^{\pm i\theta}$, and this would make the graphs appear more symmetric about $0$ than they should.  Hence, in this paper both $i$ and $-i$ would be considered to have an argument of $\frac{\pi}{2}$ radians, while a positive real number would have an argument of $0$ and a negative real number would have an argument of $\pi$.

Because calculating the eigenvalues of $J_\cI(x)$ is computationally expensive, we often work with a different matrix instead.  We define $J_\cL(x):= D (f_{enc} \circ f_{dec}) (y)= Df_{enc}(z)\cdot Df_{dec}(y)$, which is a Jacobian matrix of a map on the latent space.  If the reconstruction error is $0$ (that is, if $z=x$), the authors show in \cite{gag} that these matrices have the same nonzero eigenvalues.  In practice, we observe that when the reconstruction error is small, the two sets of eigenvalues are close to each other.  See Proposition 3.13 and Figure 2 of \cite{gag} for more details on this.

Furthermore, as we see in Theorem \ref{thm:Adhikari}, there are theoretical results for products of matrices such as $J_\cL(x)$ which do not extend to $J_\cI(x)$.  In this paper, we study the eigenvalues of $J_\cL(x)$ when the autoencoder has not been trained, and compare these to the eigenvalues of $J_\cL(x)$ when the autoencoder has been trained. Before training occurs, a weight matrix of dimensions $input\times output$ is initialized (as per pytorch's standard torch.nn.Linear) to have entries drawn independently from a uniform distribution on the interval $[-\frac{1}{\sqrt{input}},\frac{1}{\sqrt{input}}]$.  Eigenvalues of random square matrices of this type have been well-studied.  For symmetric matrices, Wigner's semicircle law of \cite{Wigner} is a classical result.  For more general matrices with independently generated entries, the circular law often provides an asymptotic description of behavior as the size of the matrix approaches infinity; a very general form of this law is due to Tao and Vu, although it was first observed by Ginibre \cite{TaoVu,Ginibre}.

Both of these results describe the limiting behavior of the {\em empirical spectral distribution} of $n\times n$ matrices $A_n$ as $n\to\infty$.

\begin{dfn} For an $n\times n$ matrix $A$, the {\bf empirical spectral distribution} of $A$ is the measure $\mu_A$ which assigns a measure of $\frac{m}{n}$ to each eigenvalue of $A$ having multiplicity $m$.  In the case when $A$ has $n$ eigenvalues of multiplicity $1$, then, this is a measure which assigns a weight of $\frac{1}{n}$ to each eigenvalue of $A$.  One of our later results will be related to the analogous empirical distribution of the squared absolute values of the eigenvalues---this distribution assigns a measure of $\frac{s}{n}$ to a real number $r$ if there are exactly $s$ eigenvalues of $A$, $\lambda_1,\dots,\lambda_s$ so that $|\lambda_1|^2=|\lambda_2|^2=\dots=|\lambda_s|^2=r$.  (See, for example, \cite{TaoVu} for a paper working with the empirical spectral distribution.)
\end{dfn}

We note, for Wigner's theorem, that a Hermitian matrix is a complex matrix $A$ so that $A=A^*$, where $A^*$ denotes the conjugate transpose of $A$ (that is, if the $(j,k)$ entry of $A$ is $a+bi$, then the $(k,j)$ entry of $A^*$ is $a-bi$).  Hermitian matrices are the complex analogue of symmetric matrices; in particular all eigenvalues of Hermitian matrices are real, and they are always diagonalizable with orthogonal eigenvectors.  Because our autoencoder's weight matrices are real matrices, for our context ``Hermitian" reduces to ``symmetric."

\begin{thm}[Wigner's semi-circle law; e.g. Theorem 1.5 of \cite{TaoVu}]\label{thm:semicircle}
    Let $A_n$ be a sequence of random $n\times n$ Hermitian matrices, where the upper diagonal entries are independent and identically distributed (i.i.d.) with mean $0$ and variance $1$.  Then the empirical spectral distribution of $\frac{1}{\sqrt{n}}A_n$ is given by the following density function on $\R$:
    \[
        \rho(x)=\begin{cases}
            \frac{1}{2\pi}\sqrt{4-x^2} & \text{when }|x|\leq2 \\
            0                          & \text{else.}
        \end{cases}
    \]
\end{thm}

Here and in the following theorems, $\frac{1}{\sqrt{n}}A_n$ denotes the matrix where the $(j,k)$ entry is $\frac{1}{\sqrt{n}}$ times the $(j,k)$ entry of $A_n$.  For instance, in the above theorem, this means that the upper diagonal entries of $\frac{1}{\sqrt{n}}A_n$ are independent, but identically distributed with a distribution that has mean $0$ and variance $\frac{1}{n}$.

We note that in the above theorem, {\em only} the upper diagonal entries can be taken to be independent, because the entry in the $(j,k)$ position is the complex conjugate of the entry in the $(k,j)$ position, due to the assumption that the matrices are Hermitian.  The next result characterizes what occurs when we allow {\em all} entries to be independent.  In this case, we do not expect our matrices to be Hermitian, and in particular we have no reason to expect their eigenvalues to be real.  Hence, this result is more applicable to the context of our autoencoders, which are not usually symmetric.

We emphasize that, because the matrices in the preceding theorem are Hermitian, the eigenvalues are all real, so the limiting distribution is a distribution on $\R$.  In the subsequent theorems, however, the matrices are more general than Hermitian or symmetric, so the distributions are distributions on $\C$ instead.

\begin{thm}[The Circular Law; Theorem 1.13 of \cite{TaoVu}]\label{thm:circular}
    Let $A_n$ be a sequence of random $n\times n$ complex matrices whose entries are i.i.d. with mean $0$ and variance $1$.  Then the empirical spectral distribution of $\frac{1}{\sqrt{n}}A_n$ almost surely converges in distribution to the uniform distribution on the unit disc.
\end{thm}

The uniform distribution having mean $0$ and variance $1$ is the uniform distribution on $[-\sqrt{3},\sqrt{3}]$.  Hence, we conclude that if a sequence of $n\times n$ matrices $A_n$ has its entries drawn from this distribution, the empirical spectral distributions of $\frac{1}{\sqrt{n}}A_n$ should converge to the uniform distribution on the unit disc.

In this case, the entries of $\frac{1}{\sqrt{n}}A_n$ are distributed uniformly in $[-\sqrt{3/n},\sqrt{3/n}]$; quite similar to the initialization for the autoencoders, which draws uniformly from $[-\sqrt{1/n},\sqrt{1/n}]$.  The difference corresponds to drawing from the distribution $[-1,1]$ rather than $[-\sqrt{3},\sqrt{3}]$ when generating the sequence $A_n$.  By renormalizing, we arrive at the following corollary.

\begin{cor}\label{cor:circle}
    If $A_n$ is a sequence of random $n\times n$ matrices with entries drawn independently from $[-1,1]$, then the empirical spectral distribution of $\frac{1}{\sqrt{n}}A_n$ almost surely converges in distribution to the uniform distribution on the disc of radius $\frac{1}{\sqrt{3}}$ about the origin.  In particular, the arguments of the eigenvalues of $\frac{1}{\sqrt{n}}A_n$ will be approximately uniformly distributed on $[-\pi,\pi]$ for sufficiently large $n$.
\end{cor}

The key intuition for this corollary is the fact that if $\lambda_1,\dots,\lambda_n$ are the eigenvalues of a matrix $A$, then the eigenvalues of the matrix $\alpha A$ (where $\alpha$ is a real or complex scalar) are $\alpha\lambda_1,\dots,\alpha\lambda_n$.  Since the analogous statement for the autoencoder distribution is to consider the eigenvalues of $\frac{1}{\sqrt{3n}}A_n$, we scale all the eigenvalues (and thus the radius of the disk) by $\frac{1}{\sqrt{3}}$.

%In particular, if we were interested in the eigenvalues of a single $input\times input$ matrix initialized from a uniform distribution on $[-\frac{1}{\sqrt{input}},\frac{1}{\sqrt{input}}]$, the circular law predicts that these eigenvalues should be uniformly distributed over a disc of radius $\frac{1}{\sqrt{3}}$ about the origin.  (The value $\sqrt{3}$ arises due to the variance of the uniform distribution on $[-1,1]$.) {\color{blue} variance of uniform on $[a, b]$ is $\frac{(b-a)^2}{12}$, so that the variance in this case is  $\frac{1}{3}$. Is the radius the variance or the st. dev.? This certainly needs to be calculated out. Make it a result of this paper.}
%\hlfix{This, in particular, would imply that the arguments of the eigenvalues would be uniformly distributed on $[-\pi,\pi)$ for sufficiently large matrices.}{make this a corrollary as you will use this in your graphs later}

We shall investigate the empirical behavior of the arguments of these eigenvalues in the next section; see in particular Figure \ref{fig:argument} to compare the distribution of the angles at initialization to the distribution after training occurs.

We note that both of these laws were initially proved for very special cases, and later generalized significantly.  The circular law in particular was first proved for the Ginibre ensembles; matrices with i.i.d. Gaussian entries \cite{Ginibre}.  As we see here, the current generalizations require only conditions on the mean and variance of the distribution generating the entries.  Hence, even though one of the results we shall discuss later in this section is stated only for the Gaussian case, we still hope that similar results for more general distributions hold (and in particular for the uniform distributions used in pytorch's torch.nn.Linear).

The previous results have focused on the case of one randomly generated matrix, when our autoencoder not only has multiple weight matrices, but also nonlinearities.  Although it is difficult to account for the effects of the nonlinearities, we wish to understand how the weight matrices should interact.  That is, we wish to understand the eigenvalues of products of random matrices.  The next two results give some insight into the expected norms of eigenvalues of such products.

\begin{dfn}
    The {\bf cumulative spectral distribution function} of an $n\times n$ matrix $A$ is the function
    \[
        F_A(x,y):=\frac{1}{n}(\#\{\text{Re}(\lambda_i)\leq x,\text{Im}(\lambda_i)\leq y\})
    \]
    where $\#\{\text{Re}(\lambda_i)\leq x,\text{Im}(\lambda_i)\leq y\}$ counts how many eigenvalues of $A$ have real part $\leq x$ and imaginary part $\leq y$ (where the counting is done with multiplicity; an eigenvalue of multiplicity $2$ counts twice).

    The {\bf expected cumulative spectral distribution function} of a random $n\times n$ matrix $A$ is simply the expected value $\mathbb{E}(F_A)$.

    %\st{(Once again, we could additionally consider the corresponding functions for the distributions of the absolute values.)}
\end{dfn}

\begin{thm}[Theorem 1.1 of \cite{GotzeTikhomirov}; rephrased for convenience]
    Let $m\geq1$ be an integer. Denote by $P_n$ the product $\frac{1}{\sqrt{n}}A_{1,n}\frac{1}{\sqrt{n}}A_{2,n}\dots\frac{1}{\sqrt{n}} A_{m,n}$ where each $A_{m,n}$ is an $n \times n$ complex matrix whose entries are i.i.d. with mean $0$ and variance $1$..  Let $F_n$ denote the expected cumulative spectral distribution function of $P_n$.  Then the sequence $F_n$ %\hlfix{converges in distribution}{Is it worth discussing different converges used in this paper in an appendix?} to the distribution whose density function on the complex plane $\C$ is given by
    converges to the distribution function $G(x,y)$ given by $G(x,y)=\int_{-\infty}^x\int_{-\infty}^yg(a+bi)\phantom{i}db\phantom{i}\;da$, where
    \[
        g(a+bi)=\begin{cases}
            \frac{1}{\pi m(a^2+b^2)^{\frac{m-1}{m}}} & \text{when }a^2+b^2\leq1 \\
            0                                        & \text{else.}
        \end{cases}
    \]
\end{thm}
The limiting distribution in the above theorem corresponds to the $m^{th}$ power of the uniform distribution on the disc; that is, the distribution which arises by choosing $z$ by the uniform distribution on the disc, and returning $z^m$.  For $m\geq1$, this distribution is hence more concentrated around $0$ than the uniform distribution.

However, this theorem applies only to products of square matrices, while the weight matrices of autoencoders are generally rectangular.  The following theorem provides one result in the rectangular case.

\begin{thm}[Theorem 11 of \cite{AdhikariEtAl}]\label{thm:Adhikari}
    Let $A_1,A_2,\dots,A_k$ be rectangular matrices with independent entries of dimensions $n_j\times n_{j+1}$ for $j=1,2,\dots,k$ with $n_{k+1}=n_1=\min\{n_1,n_2,\dots,n_k\}$, with i.i.d. entries distributed by a complex Gaussian of mean $0$ and variance $1$.  Suppose that as $n_1\to\infty$, the ratios $\frac{n_j}{n_1}\to\alpha_j$.  Then the expected empirical distribution of the squared absolute values of eigenvalues of $\frac{1}{\sqrt{n_1}}A_1\frac{1}{\sqrt{n_1}}A_2\dots\frac{1}{\sqrt{n_1}}A_k$ converges to the distribution given by
    \[
        U(U-1+\alpha_2)\dots(U-1+\alpha_k)
    \]
    where $U$ is the standard uniform distribution on $[0,1]$.
\end{thm}

Here, note that that the assumptions about the dimensions are that we move from the smallest vector space (the $n_1$) up through larger ones, then back to the smallest (since $n_1$ is the minimum).  Thus, this theorem applies exactly to the analysis of the Jacobians on the latent space, $J_\cL$, where this assumption holds.  Additionally, this theorem is a purely asymptotic statement, where the assumed regime is one where the $n_i$ increase roughly as proportions of $n_1$, $n_j\approx\alpha_j n_1$.  For the case of our autoencoder, we have $n_2,\dots,n_k$ fixed with only $n_1$ varying.

Nevertheless, we can make some quantitative predictions based on this theorem, which will be close to the empirical results we observe in the next section (in particular in Figure \ref{fig:initmodulus}). For the rest of this section, we understand the implications of Theorem \ref{thm:Adhikari} on the matrices $J_\cL(x)$ that we wish to study.

If $A_1,A_2,\dots,A_k$ are rectangular matrices with entries drawn independently from $[-1,1]$, then to imitate the initialization of our autoencoders that torch uses, we would be interested in the product matrix $\frac{1}{\sqrt{n_1}}A_1\frac{1}{\sqrt{n_2}}A_2\dots\frac{1}{\sqrt{n_k}}A_k$.  This differs from the product studied in the above theorem in two important respects.

First, rather than using a standard complex Gaussian (which has mean $0$ and variance $1$), the entries are drawn from a distribution having mean $0$ and variance $\frac{1}{\sqrt{3}}$.  Hence, we re-write the above product as
\[
    \frac{1}{\sqrt{3}^k}\Big(\frac{\sqrt{3}}{\sqrt{n_1}}A_1\frac{\sqrt{3}}{\sqrt{n_2}}A_2\dots\frac{\sqrt{3}}{\sqrt{n_k}}A_k\Big),
\]
multiplying and dividing by $\sqrt{3}$; now, for example, the matrix $\frac{\sqrt{3}}{\sqrt{n_1}}A_1$ has its entries drawn from a distribution of mean $0$ and variance $\frac{1}{n_1}$, which is closer to the context of Theorem \ref{thm:Adhikari}.

Second, in Theorem \ref{thm:Adhikari}, the matrices are all rescaled by the same factor $\frac{1}{\sqrt{n_1}}$, whereas our product uses different rescaling factors.   We fix this by multiplying and dividing by the appropriate constants, again rewriting our product as
\[
    \frac{1}{\sqrt{3}^k}\frac{\sqrt{n_1^k}}{\sqrt{n_1n_2\dots n_k}}\underbrace{\Big(\frac{\sqrt{3}}{\sqrt{n_1}}A_1\frac{\sqrt{3}}{\sqrt{n_1}}A_2\dots\frac{\sqrt{3}}{\sqrt{n_1}}A_k\Big)}_{=:P}.
\]

We now apply Theorem \ref{thm:Adhikari} to the matrix $P$ inside the parentheses, concluding that for sufficiently large $n_1$, the expected distribution of the squared absolute values of the eigenvalues of $P$ is
\[
    U(U-1+\alpha_2)\dots(U-1+\alpha_k)
\]
where $U$ is a uniform distribution on $[0,1]$ and $\alpha_j=\frac{n_j}{n_1}$.

Of course, the matrix we are actually interested in is not $P$ itself, but $\frac{1}{\sqrt{3}^k}\frac{\sqrt{n_1^k}}{\sqrt{n_1n_2\dots n_k}}$ times $P$.  Fortunately, scaling $P$ by this constant scales its eigenvalues by the same constant, and hence scales the squared absolute value of its eigenvalues by the square of this constant.  Hence, in this case the product
\[
    \frac{1}{\sqrt{n_1}}A_1\frac{1}{\sqrt{n_2}}A_2\dots\frac{1}{\sqrt{n_k}}A_k=\frac{1}{\sqrt{3}^k}\frac{\sqrt{n_1^k}}{\sqrt{n_1n_2\dots n_k}}P
\]
would have the squared absolute values of its eigenvalues distributed according to the distribution

\[
    \frac{1}{3^k\alpha_2\dots\alpha_k}U(U-1+\alpha_2)\dots(U-1+\alpha_k)
\]
which we can also rewrite as
\[
    \frac{1}{3^k}U(\frac{1}{\alpha_2}(U-1)+1)\dots(\frac{1}{\alpha_k}(U-1)+1)
\]
where $U$ is a uniform distribution on $[0,1]$ and $\alpha_j=\frac{n_j}{n_1}$.

It is plain that the minimum of this distribution is always $0$.  To compute the median and maximum, we simply substitute $0.5$ and $1$ for $U$ and compute.  These are the median and maximum of the uniform distribution, and the above expression is increasing in $U$, so we can compute medians and maxima in this simple manner.  (Note that this method would fail spectacularly for computing the mean of this distribution.)

In particular, the predicted maximum is quite easy to compute, as it is independent of the $\alpha_i$; it is simply $\frac{1}{3^k}$.  The median is more involved.

To imitate the architecture of our autoencoder, we set $k=8$ and $n_2=n_8=32$, $n_3=n_7=64$, $n_4=n_6=128$, and $n_5=784$.  In this case, the predicted median squared norm of the eigenvalues is given by
\begin{align*}
    \frac{1}{3^k}(0.5)(\frac{n_1}{32}(-0.5)+1)^2(\frac{n_1}{64}(-0.5)+1)^2 \\\times(\frac{n_1}{128}(-0.5)+1)^2(\frac{n_1}{784}(-0.5)+1).
\end{align*}

For $n_1=2$, this expression is about $0.00013$, and for $n_1=20$, it is about $0.000043$.  It decreases slowly in $n_1$ in the range $[2,20]$; the corresponding maximum is constant at $\frac{1}{3^8}\approx0.00015$.  The implied median eigenvalue norm (not squared) is then about $0.011$ for $n_1=2$, and about $0.0066$ for $n_1=20$; the implied maximum eigenvalue norm is $\frac{1}{81}\approx0.012$.

In the Experiment section, we will compare these predictions to the observations shown in Figure \ref{fig:initmodulus}.

Finally, we remark that the last theorem applied to matrices with entries drawn from a standard complex Gaussian.  Because a complex Gaussian is rotation-invariant about $0$, the predicted distribution of eigenvalues of such matrices is also rotation-invariant; for this reason the last theorem makes no comment on the distribution of arguments.  While there is solid theoretical work (e.g. \cite{AkemannIpsen,AkemannBurda,IpsenKieburg}) on the eigenvalues of products of matrices with entries drawn from real Gaussians, this work does not lend itself to a quickly explicable distribution of the arguments of these eigenvalues.  Hence, the best theoretical tool we have for the arguments of the eigenvalues is Theorem \ref{thm:circular} and its Corollary \ref{cor:circle}, which correspond to a uniform distribution of angles.%Hence, even without accounting for the nonlinearities of the autoencoder's structure, it is difficult to make principled guesses about the behavior of the arguments before training occurs. \sanote{rephrase this more positively. you are talking to a more experimental audience than most mathematicians}

\section{Experiment}

We trained nineteen autoencoders on the standard training points of the MNIST dataset.  These have identical architecture, except for the dimension of the latent space $\cL$.  The encoder and decoder each have four layers, with ReLU activation functions for all layers.  The encoder layers are $(\R^{784},\R^{128},\R^{64},\R^{32},\R^d=\cL)$ and the decoder layers are $(\cL=\R^d,\R^{32},\R^{64},\R^{128},\R^{784})$, where $d\in\{2,3,\dots,20\}$ are the latent dimensions.

The input and reconstruction spaces for these autoencoders are $\R^{784}$, because the MNIST images are $28\times28$, and are flattened into a vector for the training.  In addition to the ReLU activation functions for the layers, the decoder has a component wise $\tanh$ function at the end, which normalizes the outputs to be in $[-1,1]^{784}$. We also scale the input space to be in $[-1,1]^{784}$.

We trained these autoencoders for 300 epochs with seed $0$.  We show observations at 0, 1, 4, 10, and 50 epochs as well.

\subsection{Eigenvalue behavior}

\begin{figure*}[h!] 
    \centering
    \begin{subfigure}[t]{0.3\linewidth}
        \includegraphics[width=\textwidth]{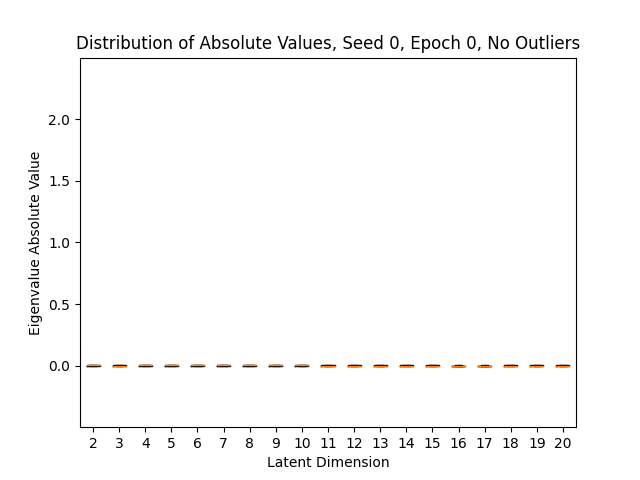}
    \end{subfigure}
    \begin{subfigure}[t]{0.3\linewidth}
        \includegraphics[width=\textwidth]{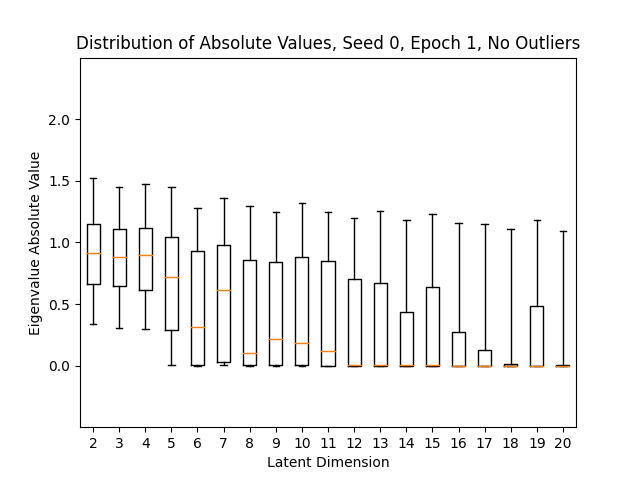}
    \end{subfigure}
    \begin{subfigure}[t]{0.3\linewidth}
        \includegraphics[width=\textwidth]{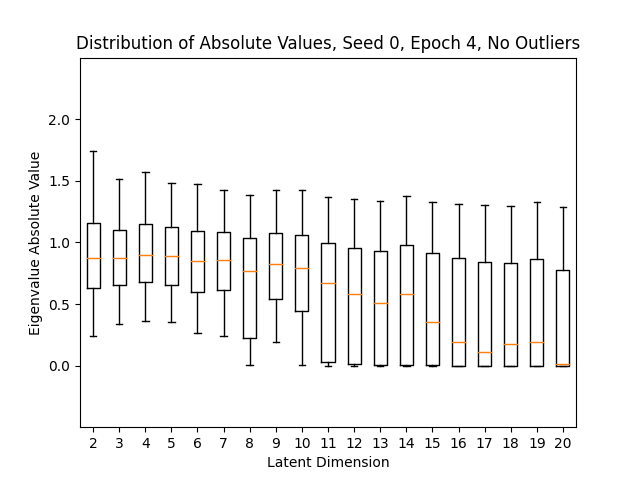}
    \end{subfigure}
\par
    \begin{subfigure}[t]{0.3\linewidth}
        \includegraphics[width=\textwidth]{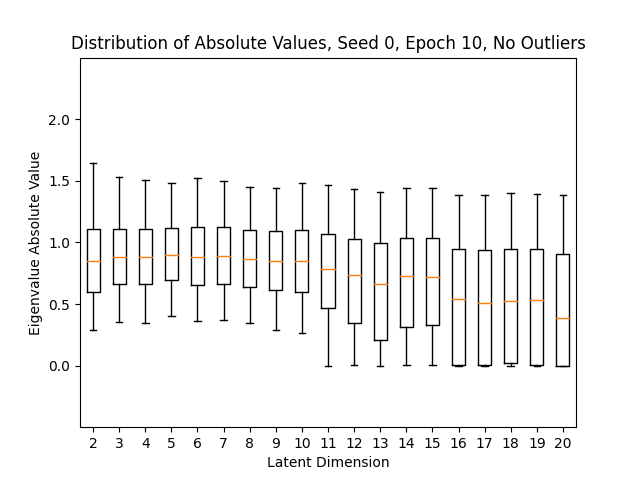}
    \end{subfigure}
    \begin{subfigure}[t]{0.3\linewidth}
        \includegraphics[width=\textwidth]{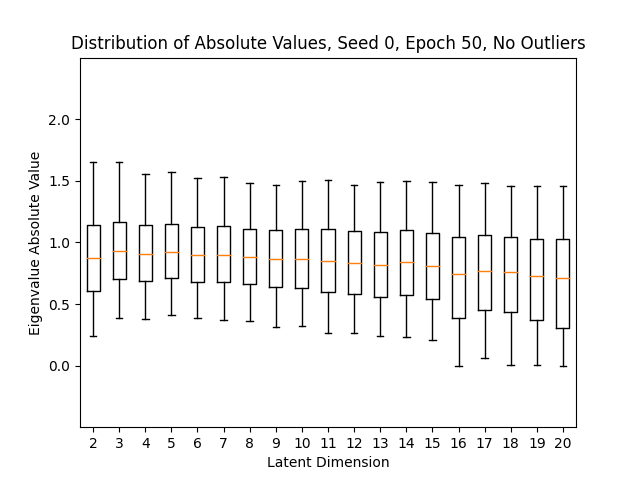}
    \end{subfigure}
    \begin{subfigure}[t]{0.3\linewidth}
        \includegraphics[width=\textwidth]{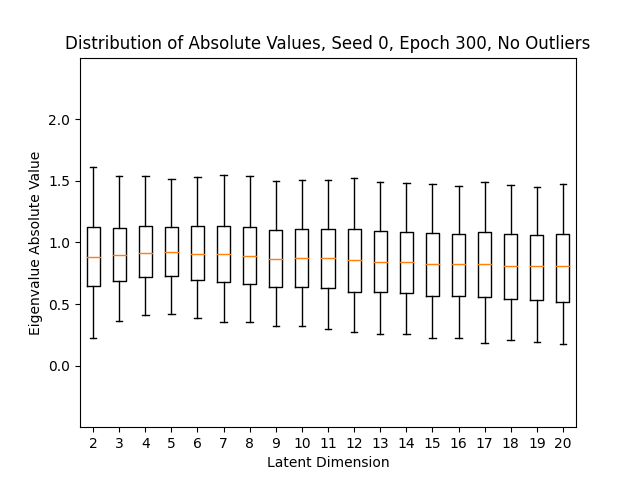}
    \end{subfigure}
    \caption{The box plots for $|\lambda|$, where $\lambda$ are eigenvalues of $J_\cL(x)$ for $x$ an MNIST data point. From left to right, top to bottom graphs give the moduli at initialization (epoch $0$), epoch $1$, epoch $4$, epoch $10$, epoch $50$ and epoch $300$ (the fully trained model).  As all graphs are on the same scale, epoch $0$ is reproduced in Figure \ref{fig:initmodulus}. Note that as training progresses, the distribution of $|\lambda|$ becomes more consistent across architecture converging to a median of $1$.}
    \label{fig:modulus}
\end{figure*}

In \cite{gag}, the authors observe that the eigenvalues of the fully trained autoencoders display some properties that we would expect if $f_{dec}\circ f_{enc}$ were a projection onto a data manifold.  In particular, when we consider the eigenvalues of $J_\cI(x)$ or $J_\cL(z)$ for $x$ in the MNIST dataset, the norms of the eigenvalues are usually less than $1$.  Moreover, as the latent dimension increases, the proportion of eigenvalues with small norms increases.  There are not as many eigenvalues with small norms as we would expect from projections, however.  Additionally, the arguments of the eigenvalues (the angles with the positive real axis---if we write an eigenvalue $\lambda=re^{i\theta}$ in polar, the argument is the angle $|\theta|$) tend to be small.  In particular, there are very few negative eigenvalues, and relatively few complex eigenvalues with negative real parts.  See the bottom right graphs of Figure \ref{fig:modulus} and Figure \ref{fig:argument} for the distributions of the norms and arguments of the eigenvalues of the fully trained autoencoders (300 epochs). %and relatively few datapoints where the product of the eigenvalues (the determinant of $J_\cL$) is negative.  Geometrically, this means that the autoencoder does not reverse the orientation of the latent space near the data points. \sanote{some of these results need graphs}

At initialization, the eigenvalues of these matrices are determined by the process chosen to initialize the autoencoders.  As mentioned previously, the preceding section's results do not account for the nonlinearities of the autoencoder's architecture, but do describe how we would expect the weight matrices to interact without those nonlinearities.
%The theoretical results of the preceding section do not apply precisely, as none of them account for the nonlinearities of the autoencoder's architecture.  \sanote{Do they describe the emipircal results in some qualitative way instead?} 

We first consider the absolute values of the eigenvalues.  Based on Theorem \ref{thm:Adhikari} and the subsequent computations, we expect the eigenvalues to be relatively small (with medians ranging from around $0.01$ in latent dimension $2$ to $0.006$ in latent dimension $20$).  Qualitatively, we expect the median eigenvalue to decrease as the latent dimension increases from $2$ to $20$.

\begin{figure}
    \centering
    \includegraphics[width=0.5\textwidth]{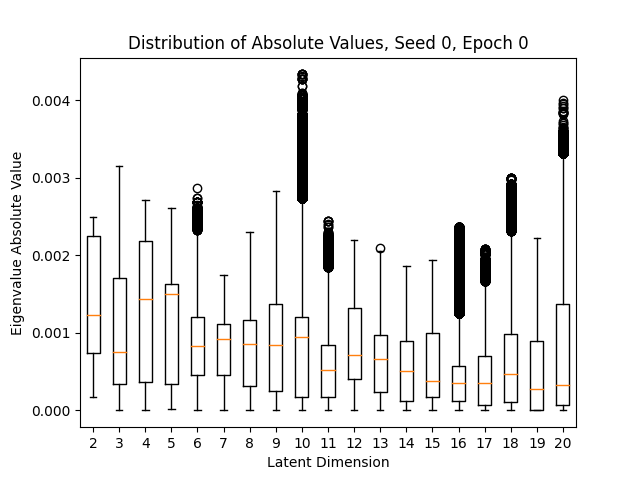}
    \caption{The box plots for $|\lambda|$ at epoch $0$.  This graph is at a different scale than the graphs of Figure \ref{fig:modulus}.  We also include the outliers, so that the maximum eigenvalue norm can be observed. All of the eigenvalues are small, and the median norm decreases wtih latent dimension.}
    \label{fig:initmodulus}
\end{figure}

\begin{figure*}[h!]
    \centering
    \begin{subfigure}[]{0.3\linewidth}
        \includegraphics[width=\textwidth]{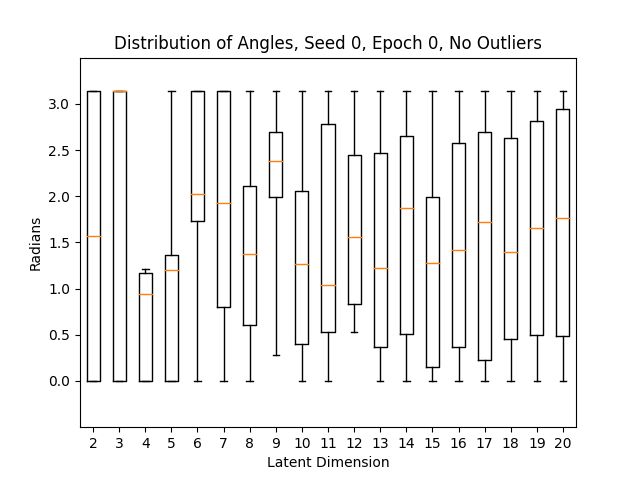}
    \end{subfigure}
    \begin{subfigure}[]{0.3\linewidth}
        \includegraphics[width=\textwidth]{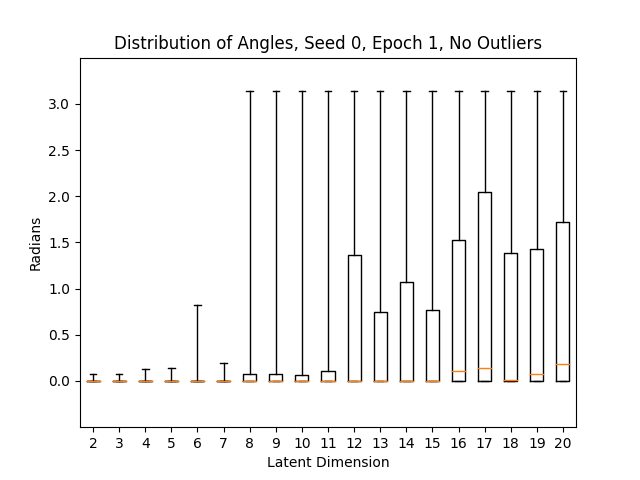}
    \end{subfigure}
    \begin{subfigure}[]{0.3\linewidth}
        \includegraphics[width=\textwidth]{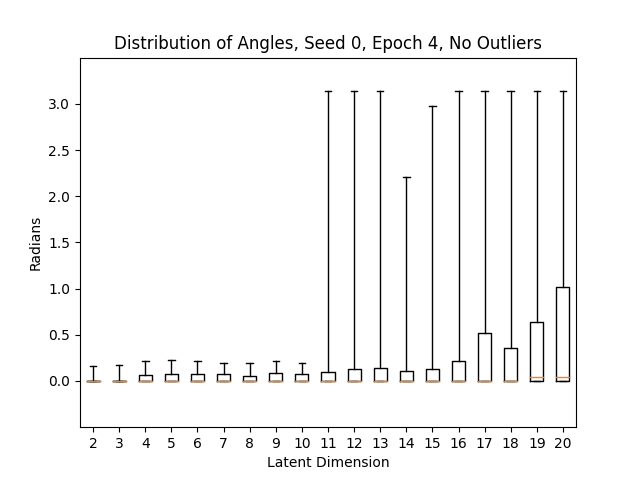}
    \end{subfigure}
\par
    \begin{subfigure}[]{0.3\linewidth}
        \includegraphics[width=\textwidth]{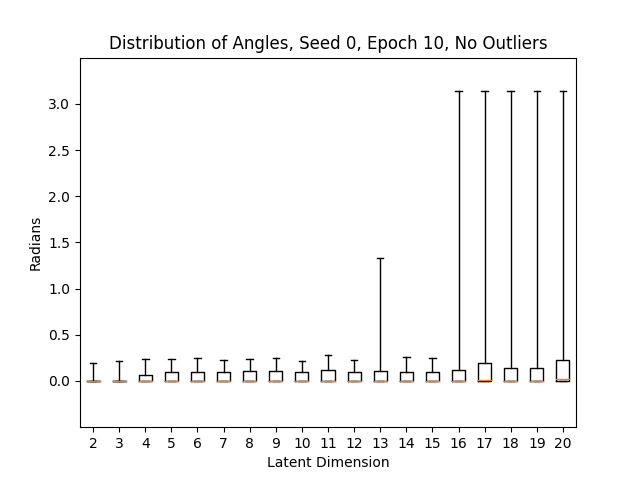}
    \end{subfigure}
    \begin{subfigure}[]{0.3\linewidth}
        \includegraphics[width=\textwidth]{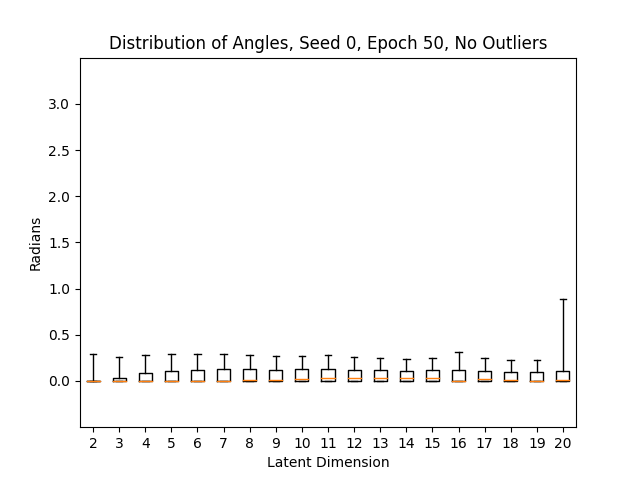}
    \end{subfigure}
    \begin{subfigure}[]{0.3\linewidth}
        \includegraphics[width=\textwidth]{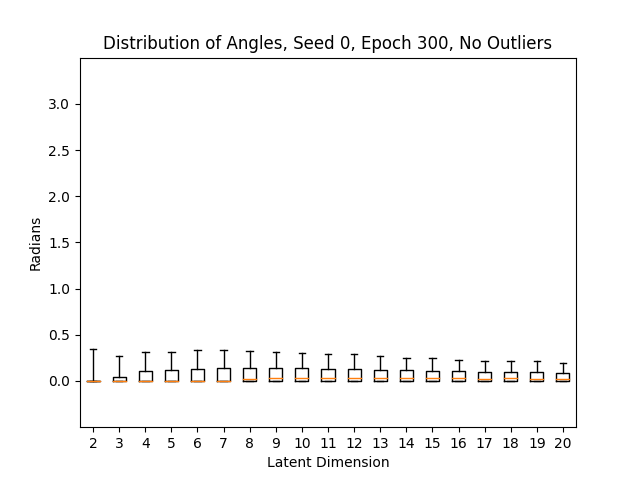}
    \end{subfigure}
    \caption{The box plots for the arguments of the eigenvalues of $J_\cL(x)$, for $x$ in the MNIST dataset.  From left to right, top to bottom graphs give the moduli at initialization (epoch $0$), epoch $1$, epoch $4$, epoch $10$, epoch $50$ and epoch $300$ (the fully trained model). Note that as training progresses, the arguments become small.}
    \label{fig:argument}
\end{figure*}

In fact, we observe in Figure \ref{fig:initmodulus} that all eigenvalues at epoch $0$ have extremely small norms---much smaller than the eigenvalues of the trained autoencoders, at most on the order of $0.005$ (where the fully trained autoencoders have most of their eigenvalues between $0.2$ and $1.5$).  Indeed, these values are slightly smaller than the predicted values, but are generally within an order of magnitude of the theoretical predictions. Overall, the median eigenvalue does appear to decrease as the latent dimensions increase, matching the qualitative prediction.  Hence, although Theorem \ref{thm:Adhikari} considers only the products of random matrices, it still offers useful predictions for the eigenvalues of autoencoders, despite the nonlinearities of their architectures.

We note that with training, these median eigenvalues quickly converge to $1$, doing so faster for the autoencoders with lower latent dimension (which are easier to train, as they are learning fewer features) than those with higher latent dimension.

Our expectations for the arguments of the eigenvalues are based on Corollary \ref{cor:circle}, which predicts that the arguments should be uniformly distributed on $[-\pi,\pi]$, for large enough latent dimensions.  Because we consider the absolute value of the arguments, this suggests that we should see a uniform distribution on $[0,\pi]$.  We expect the median argument, then, to be $\frac{\pi}{2}$; about $1.57$ radians.

We graph the observed arguments at epoch 0 in the top left chart of Figure \ref{fig:argument}.  We observe that the median arguments display considerable variation, as does the distribution as a whole.  However, as the latent dimension increases, the median arguments do tend towards $\frac{\pi}{2}$, and the distribution becomes more uniform, consistent with the theoretical predictions.  Overall, at initialization, the distributions of the eigenvalues bear little resemblance to the fully trained model and are instead better described by theoretical results pertaining to eigenvalues of random matrices, as one might expect.

As the training progresses, the eigenvalues begin to display the qualitative features of the fully trained autoencoders.  Even at epoch $1$, in low dimensions the median eigenvalue is far from $0$, and the distribution has clearly shifted from the initialization.  By epoch $10$, the median eigenvalue modulus is at least $0.5$ in all latent dimensions; by epoch $50$, this median is near $1$ in all latent dimensions, and the distributions quite closely resemble those of the fully trained autoencoder.

Similarly, at epoch $1$ the arguments of the eigenvalues are clearly not uniformly distributed for any latent dimensions. By epoch $10$, the median argument of the eigenvalues is close to $0$, consistent with the fully trained autoencoder.  The 95th percentile of the arguments remains fairly high through epoch $50$, at least in the largest latent dimensions.  However, we do not see many negative eigenvalues by epoch $50$ (although in high latent dimensions, more than $5\%$ of the eigenvalues are still negative in epoch $10$).  In the fully trained autoencoder, the vast majority of the eigenvalue arguments are quite close to $0$.

Overall, the norms of the eigenvalues seem to train more quickly than the arguments.  Although both change very quickly away from the initialized distribution, there are still many extremely large angles in epoch $10$ and even in epoch $50$, the highest latent dimension still has a strikingly large 95$^{\text{th}}$ percentile.  However, we do clearly see that the trained behavior of the autoencoders' eigenvalues emerges during the training process, particularly since the arguments shink faster for the autoencoders with lower latent dimension (which are easier to train, as they are learning fewer features) than those with higher latent dimension.

\section{Connection to other work}

Randomness has been used in machine learning algorithms in many ways. For instance, Random Vector Functional Link architectures have been proposed for simple networks \cite{PaoTakefuji} and have been shown to be successful at learning the manifold structure of the data over the subsequent decades \cite{NeedellEtAl, HuangEtAl, KatuwalEtAl}. These methods were initially developed to increase the speed of training a network.

Random matrices are used in certain backpropagation techniques, such as feedback alignment \cite{LillicrapEtAl, Nokland}, random backpropagation \cite{BaldiEtAl}, and related algorithms \cite{FrenkelEtAl, HanEtAl}. In this work, the transpose of the forward weight matrices in the back propagation process is replaced with a random matrix, increasing training speed, generalizability and mimicking biological processes.  In our paper, on the other hand, we perform a more standard backpropagation, and are interested in random matrices because these are the starting point for this training process.

The work in this paper is about understanding the geometric evolution of the learning process through a standard feedforward network and standard backpropagation. The eigenvalues of the function learned by a neural network give information on the structure of the feature map on a dataset. In this respect, the work is similar to work studying the eigenvalues of Hessian matrices of the loss, such as \cite{SagunEtAl1,SagunEtAl2, GhorbaniEtAl}.  Indeed, related work also reveals the importance of the early phases of training; see in particular \cite{FrankleEtAl} and \cite{GurAriEtAl}.  These works show that early on in training, the gradient descent converges to a small subspace, spanned by the top eigenvectors of the Hessian.  Our work also shows the importance of the early epochs, as the autoencoders' eigenvalue distributions rapidly approach those of a fully trained model in this period.  However, while both these works and ours focus on using eigenvalues to draw geometric conclusions, they focus on the geometry of the training space, as described by the eigenvectors of the Hessian, while we hope to understand the geometry of the data manifold itself with the eigenvalues of the autoencoder.

\section{Conclusion}

This paper investigates the eigenvalues of an autoencoder's Jacobian matrices, and how these change during training.  Understanding how these eigenvalues evolve gives insight into the geometric properties of the feature map, and how they evolve through training. There is a rich body of mathematical results on the eigenvalues of a large family of random matrices. We have seen that these results qualitatively describe the distribution of the eigenvalues at initialization. In particular, the eigenvalues are small in absolute value and, at least in the high latent dimensions, display a distribution of angles that is close to uniform.

After initialization, we see that these distributions rapidly evolve to resemble those of a trained autoencoder, with eigenvalue norms becoming much larger (on the order of $0.5$ to $1$, rather than $0.001$ to $0.005$), and with angles becoming concentrated close to $0$.  Qualitatively, this supports the intuition that a trained autoencoder begins to resemble a projection onto a data manifold, as the authors previously observed in \cite{gag}.  This paper additionally shows that this pattern is in fact a result of the training, as it is quite plainly not present at initialization.

\bibliographystyle{amsplain}
\bibliography{AutoEncoderInitial}

\providecommand{\bysame}{\leavevmode\hbox to3em{\hrulefill}\thinspace}
\providecommand{\MR}{\relax\ifhmode\unskip\space\fi MR }
\providecommand{\MRhref}[2]{%
  \href{http://www.ams.org/mathscinet-getitem?mr=#1}{#2}
}
\providecommand{\href}[2]{#2}
\begin{thebibliography}{10}

\bibitem{AdhikariEtAl}
Kartick {Adhikari}, Nanda~Kishore {Reddy}, Tulasi~Ram {Reddy}, and Koushik
  {Saha}, \emph{{Determinantal point processes in the plane from products of
  random matrices}}, Annales de L'Institut Henri Poincare Section (B)
  Probability and Statistics \textbf{52} (2016), no.~1, 16--46.

\bibitem{Peterson}
R.~Beveridge adn Jen-Mei~Chang, B.~Draper, M.~Kirby, H.~Kley, and C.~Peterson,
  \emph{Principal angles separate subject illumination spaces in {YDB} and
  {CMU-PIE}}, {IEEE} Trans. Pattern Analysis and Machine Intelligence
  \textbf{31} (2009), no.~2, 351--363.

\bibitem{gag}
S.~Agarwala, B.~Dees, A.~S. Gearheart, and C.~Lowman, \emph{Geometry and
  generalization: Eigenvalues as predictors of where a network will fail to
  generalize}, arXiv:2107.06386.

\bibitem{AkemannBurda}
Gernot Akemann and Zdzislaw Burda, \emph{Universal microscopic correlation
  functions for products of independent ginibre matrices}, Journal of Physics
  A: Mathematical and Theoretical \textbf{45} (2012), no.~46, 465201.

\bibitem{AkemannIpsen}
Gernot Akemann and Jesper Ipsen, \emph{Recent exact and asymptotic results for
  products of independent random matrices}, Acta Physica Polonica B \textbf{46}
  (2015).

\bibitem{ID}
A.~Ansuini, A.~Laio, J.H. Macke, and D.~Zoccolan, \emph{Intrinsic dimension of
  data representations in deep neural networks}, Advances in Neural Information
  Processing Systems (2019), 6111--6122.

\bibitem{BaldiEtAl}
Pierre Baldi, Peter Sadowski, and Zhiqin Lu, \emph{Learning in the machine},
  Artif. Intell. \textbf{260} (2018), no.~C, 1--35.

\bibitem{EvansRosenthal}
M.J. Evans and J.S. Rosenthal, \emph{Probability and statistics: The science of
  uncertainty}, W. H. Freeman, 2004.

\bibitem{moreID}
E.~Facco, M.~d’Errico, A.~Rodriguez, and A.~Laio, \emph{Estimating the
  intrinsic dimension of datasets by a minimal neighborhood information},
  Scientific reports \textbf{7} (2017), no.~1, 1--8.

\bibitem{FrankleEtAl}
Jonathan Frankle, David~J. Schwab, and Ari~S. Morcos, \emph{The early phase of
  neural network training}, International Conference on Learning
  Representations, 2020.

\bibitem{FrenkelEtAl}
Charlotte Frenkel, Martin Lefebvre, and David Bol, \emph{Learning without
  feedback: Fixed random learning signals allow for feedforward training of
  deep neural networks}, Frontiers in Neuroscience \textbf{15} (2021).

\bibitem{GhorbaniEtAl}
Behrooz Ghorbani, Shankar Krishnan, and Ying Xiao, \emph{An investigation into
  neural net optimization via hessian eigenvalue density}, Proceedings of the
  36th International Conference on Machine Learning (Kamalika Chaudhuri and
  Ruslan Salakhutdinov, eds.), Proceedings of Machine Learning Research,
  vol.~97, PMLR, 09--15 Jun 2019, pp.~2232--2241.

\bibitem{Ginibre}
Jean {Ginibre}, \emph{{Statistical Ensembles of Complex, Quaternion, and Real
  Matrices}}, Journal of Mathematical Physics \textbf{6} (1965), no.~3,
  440--449.

\bibitem{GotzeTikhomirov}
Friedrich Goetze and A.~Tikhomirov, \emph{On the asymptotic spectrum of
  products of independent random matrices},  (2010).

\bibitem{GurAriEtAl}
Guy Gur-Ari, Daniel~A. Roberts, and Ethan Dyer, \emph{Gradient descent happens
  in a tiny subspace}, 2018.

\bibitem{HanEtAl}
Donghyeon Han, Jinsu Lee, Jinmook Lee, and Hoi-Jun Yoo, \emph{A 1.32 tops/w
  energy efficient deep neural network learning processor with direct feedback
  alignment based heterogeneous core architecture}, 2019 Symposium on VLSI
  Circuits, 2019, pp.~C304--C305.

\bibitem{HuangEtAl}
Guang-Bin Huang, Qin-Yu Zhu, and Chee-Kheong Siew, \emph{Extreme learning
  machine: Theory and applications}, Neurocomputing \textbf{70} (2006), no.~1,
  489--501, Neural Networks.

\bibitem{IpsenKieburg}
Jesper~R. Ipsen and Mario Kieburg, \emph{Weak commutation relations and
  eigenvalue statistics for products of rectangular random matrices}, Physical
  Review E \textbf{89} (2014), no.~3.

\bibitem{LillicrapEtAl}
Timothy~P. Lillicrap, Daniel Cownden, Douglas~B. Tweed, and Colin~J. Akerman,
  \emph{Random synaptic feedback weights support error backpropagation for deep
  learning}, Nature Communications \textbf{7} (2016), no.~1, 13276.

\bibitem{UMAP}
L.~McInnes, J.~Healy, N.~Saul, and L.~Grossberger, \emph{{UMAP}: Uniform
  manifold approximation and projection}, Journal of Open Source Software
  \textbf{3} (2018), no.~29, 861.

\bibitem{NeedellEtAl}
Deanna Needell, Aaron~A. Nelson, Rayan Saab, and Palina Salanevich,
  \emph{Random vector functional link networks for function approximation on
  manifolds}, 2020.

\bibitem{Nokland}
Arild N\o~kland, \emph{Direct feedback alignment provides learning in deep
  neural networks}, Advances in Neural Information Processing Systems (D.~Lee,
  M.~Sugiyama, U.~Luxburg, I.~Guyon, and R.~Garnett, eds.), vol.~29, Curran
  Associates, Inc., 2016.

\bibitem{PaoTakefuji}
Y.-H. Pao and Y.~Takefuji, \emph{Functional-link net computing: theory, system
  architecture, and functionalities}, Computer \textbf{25} (1992), no.~5,
  76--79.

\bibitem{SagunEtAl1}
Levent Sagun, Leon Bottou, and Yann LeCun, \emph{Eigenvalues of the hessian in
  deep learning: Singularity and beyond}, 2017.

\bibitem{SagunEtAl2}
Levent Sagun, Utku Evci, V.~Ugur Guney, Yann Dauphin, and Leon Bottou,
  \emph{Empirical analysis of the hessian of over-parametrized neural
  networks}, 2018.

\bibitem{KatuwalEtAl}
Qiushi Shi, Rakesh Katuwal, Ponnuthurai Suganthan, and M.~Tanveer, \emph{Random
  vector functional link neural network based ensemble deep learning}, Pattern
  Recognition \textbf{117} (2021), 107978.

\bibitem{TaoVu}
Terence Tao and Van Vu, \emph{Random matrices: Universality of esds and the
  circular law}, The Annals of Probability \textbf{38} (2010), no.~5,
  2023--2065.

\bibitem{tSNE}
L.~van~der Maaten, \emph{Learning a parametric embedding by preserving local
  structure}, Artificial Intelligence and Statistics (2009), 384--391.

\bibitem{Wigner}
Eugene~P. Wigner, \emph{On the distribution of the roots of certain symmetric
  matrices}, Annals of Mathematics \textbf{67} (1958), no.~2, 325--327.

\bibitem{xu2021how}
Keyulu Xu, Mozhi Zhang, Jingling Li, Simon~Shaolei Du, Ken-Ichi Kawarabayashi,
  and Stefanie Jegelka, \emph{How neural networks extrapolate: From feedforward
  to graph neural networks}, International Conference on Learning
  Representations, 2021.

\end{thebibliography}
\clearpage

\section{Appendix}
\subsection{Autoencoder Structure and Jacobians}
For the autoencoders considered here, we have $f_{enc}=f_{enc,4}\circ f_{enc,3}\circ f_{enc,2}\circ f_{enc,1}$ where each $f_{enc,i}$ is a composition of a linear map with an activation function (except $f_{enc,4}$ which omits the activation function), and $f_{dec}=f_{dec,4}\circ f_{dec,3}\circ f_{dec,2}\circ f_{dec,1}$.  In the following, $A_{n\times m}$ denotes a weight matrix of dimensions $n\times m$, $b_n$ denotes a bias vector $n$ entries long, and $ReLU_n$ and $\tanh_n$ denote the functions which take in a vector of length $n$ and apply the function $ReLU$ or $\tanh$ (respectively) to each entry of that vector.  The maps comprising the encoder function have the form:
\begin{align*}
    f_{enc,1} & =ReLU_{128}(A_{128\times784}x+b_{128}) \\
    f_{enc,2} & =ReLU_{64}(A_{64\times128}x+b_{64})    \\
    f_{enc,3} & =ReLU_{32}(A_{32\times64}x+b_{32})     \\
    f_{enc,4} & =A_{d\times 32}x+b_d.
\end{align*}
Similarly, the decoder maps have the form:
\begin{align*}
    f_{dec,1} & =ReLU_{32}(A_{32\times d}x+b_{32})       \\
    f_{dec,2} & =ReLU_{64}(A_{64\times32}x+b_{64})       \\
    f_{dec,3} & =ReLU_{128}(A_{128\times64}x+b_{128})    \\
    f_{dec,4} & =\tanh_{784}(A_{784\times128}x+b_{784}).
\end{align*}

In this, $d$ represents the latent dimension, which ranges from $2$ to $20$.

To compute the Jacobian matrix $J_\cL$ or $J_\cI$, we repeatedly apply the chain rule.  Hence, to compute these Jacobians it suffices to know the derivatives of the functions comprising the autoencoder.  The derivative of an affine linear map $A_{n\times m}x + b_n$ is simply the matrix $A_{n\times m}$.  The derivative of the $ReLU_n$ map is the Heaviside function $H(x)$ on the diagonals of an $n\times n$ matrix, where the Heaviside function is
\[
    H(x)=\begin{cases}1 & x>0\\0 & x\leq0.\end{cases}
\]
This derivative matrix at a point $(x_1,\dots,x_n)$ is the diagonal matrix where the $(j,j)$ entry is $1$ when $x_j>0$, and is $0$ otherwise.  Similarly, the derivative of $\tanh_n$ is $1-\tanh^2(x)$ on the diagonal of an $n\times n$ matrix.

\subsection{Various Forms of Probabilistic Convergence}
This section focuses on the forms of convergence that appear in the cited theorems.  For the omitted proofs, and for additional context, we refer the interested reader to \cite{EvansRosenthal}.

It is usual in the theory of probability to begin with some ``sample space" $\Omega$, which represents the possible events under consideration.  We also consider a suitably large class $\mathcal{M}$ of ``measurable subsets" of $\Omega$, which are the sets to which we may assign a probability.  Finally, we have a probability function $\text{Pr}:\mathcal{M}\to[0,1]$, satisfying a number of properties which encode the idea of being a probability,
\begin{enumerate}
    \item $\text{Pr}(\Omega)=1$
    \item For $A,B\in\mathcal{M}$ with $A\cap B=\varnothing$, $\text{Pr}(A\cup B)=\text{Pr}(A)+\text{Pr}(B)$
    \item For $A_1,A_2,\dots\in\mathcal{M}$ so that $A_i\cap A_j=\varnothing$ for all $i,j$,
          \[
              \text{Pr}\Big(\bigcup_{i=1}^\infty A_i\Big)=\sum_{i=1}^\infty \text{Pr}(A_i)
          \]
    \item For each $A\in\mathcal{M}$, $\text{Pr}\geq0$.
\end{enumerate}

For example, we could take $\Omega=[0,1]$, $\mathcal{M}$ to be the Lebesgue measurable sets of $[0,1]$, and $\text{Pr}$ to be the measure of a set, which we think of as its length.  (In particular, for an interval $[a,b]\subset[0,1]$, the measure is literally equal to $b-a$, the length of the interval.)

Another common example is to take $\Omega=\R$, $\mathcal{M}$ to be the Lebesgue measurable sets of $\R$, and $\text{Pr}$ to correspond to a Gaussian.  In this case, for any measurable set $A\subset\R$, we set
\[
    \text{Pr}(A)=\frac{1}{\sqrt{2\pi}}\int_Ae^{-x^2/2}dx.
\]

We represent random variables as functions from our sample space $\Omega$.  For example, we can consider random real numbers by considering functions $f:\Omega\to\R$, we can consider random complex variables by considering functions $f:\Omega\to\C$, and so on.  Indeed, we can even consider random {\em random distributions} by considering functions whose outputs are probability distributions.  (For example, the empirical spectral distribution of a random matrix is precisely such a random variable.)

Two of the theorems cited above deal with the expectation or expected values of a random variable.  This is a notion intuitively tied to the idea of an average value or a weighted average; the formal measure-theoretic definition is \[\mathbb{E}(X):=\int_\Omega X(\omega) d\text{Pr}(\omega)\]
where the integral on the right-hand side of the definition is a Lebesgue integral with respect to the measure $\text{Pr}$.  We remark that this applies more generally than when $X$ is a real random variable---the formula defining $\mathbb{E}(X)$ makes sense as long as a suitable form of the integral can be defined.

\begin{dfn}
    We say that a sequence of random vectors $X_1,X_2,\dots\in\R^k$ {\bf converges in distribution} to the random vector $X$ if
    \[
        \lim_{n\to\infty}\text{Pr}(X_n\in A)=\text{Pr}(X\in A)
    \]
    for all measurable sets $A$ such that $\text{Pr}(X\in\partial A)=0$ (where $\partial A$ denotes the boundary of $A$; e.g. the boundary of the unit disc $\{x^2+y^2\leq 1\}$ is the unit circle $\{x^2+y^2=1\}$).
\end{dfn}

For random real variables, this can be more conveniently stated in terms of the cumulative distribution functions of the variables; if we let $F_n(x)=\text{Pr}(X_n\leq x)$ and $F(x)=\text{Pr}(X\leq x)$ be these cumulative distribution functions, an equivalent definition is that, for every $x$ at which $F(x)$ is continuous, we have
\[
    \lim_{n\to\infty}F_n(x)=F(x).
\]

This is a relatively mild form of convergence; it is in particular implied by the other two forms of convergence we shall discuss.  For proofs of these implications, see Section 4.7 of \cite{EvansRosenthal}.

\begin{dfn}
    We say that a sequence of random vectors $X_1,X_2,\dots\in\R^k$ {\bf converges in probability} to the random vector $X$ if, for every $\epsilon>0$,
    \[
        \lim_{n\to\infty}\text{Pr}(|X_n-X|>\epsilon)=0.
    \]

    That is, as we go further and further out in our sequence, the probability of $X_n$ being far from $X$ decreases to $0$.  Additionally, this definition can be applied not only to random vectors, but to random variables valued in any metric space (i.e. a set equipped with a notion of distance, which allows us to say when points are close together and far apart).
\end{dfn}

This implies convergence in distribution, and is in turn implied by the third mode of convergence we discuss (again, see Section 4.7 of \cite{EvansRosenthal} for details).

\begin{dfn}
    We say that a sequence of random variables $X_1,X_2,\dots$ {\bf converges almost surely} to the random variable $X$ if
    \[
        \text{Pr}(\lim_{n\to\infty}X_n=X)=1.
    \]

    That is, if we pick an $\omega\in\Omega$ and consider the values $X_1(\omega),X_2(\omega),\dots$, this sequence converges to $X(\omega)$ with probability $1$.
\end{dfn}

We now discuss the phrase \emph{converges almost surely in distribution} that appears in the statements of the circle law (Theorem \ref{thm:circular}) and the semi-circle law (Theorem \ref{thm:semicircle}). In this case, it is important to recall that we are dealing with random \emph{random functions}. That is, given an input $\omega \in \Omega$, the function $\mu_i$ whose convergence properties we are interested in gives a probability distribution. Then, with probability $1$, if we pick $\omega \in \Omega$, the sequence of probability distributions $\mu_1(\omega), \mu_2(\omega) \ldots, $ converges (in distribution) to the probability distribution $\mu(\omega)$.

%The context for this is when we have a sequence of random probability distributions $\mu_1,\mu_2,\dots$ (which are functions from $\Omega$, our sample space---here it is most helpful to think of $\Omega$ as being the space of random matrices).  Suppose that we claim that $\mu_1,\mu_2,\dots$ converges almost surely in distribution to $\mu$.  The ``almost surely" of this phrase means that if we pick some $\omega\in\Omega$, with probability $1$ we will pick an $\omega$ so that the sequence of distributions $\mu_1(\omega),\mu_2(\omega),\dots$ converges in distribution to $\mu(\omega)$.

It is then perhaps most helpful to unpack the ``converges in distribution" half of the definition by way of an example.  If, for instance, $\mu$ is the uniform distribution on the unit disc, to say that $\mu_1,\mu_2,\dots$ converges in distribution to $\mu$ means firstly that
\[
    \lim_{n\to\infty}\text{Pr}(|\mu_n|>1)=0.
\]

That is, the proportion of $\mu_n$ that lies outside of the unit disc must approach $0$ as $n\to\infty$.  Secondly, if $A$ is a subset of the unit disc, and the area of $A$ is $\alpha$ times the area of the unit disc, then
\[
    \lim_{n\to\infty}\text{Pr}(\mu_n\in A)=\alpha.
\]
More concretely, if we take a half disc $H$, this says that for large enough $n$, about half of $\mu_n$ must be in $H$.  Visually speaking, this says that the measures $\mu_n$ cannot cluster in any part of the unit disc; rather they must begin to appear uniform.
%While we do not demonstrate here that this implies convergence in probability, the following example demonstrates that convergence in probability does not imply almost sure convergence.
%By $1_{[a,b]}$ we denote the function \[1_{[a,b]}(x)=\begin{cases}1&x\in[a,b]\\0&\text{else.}\]
%Consider the sequence $1_{[0,1]},1_{[0,1/2]},1_{[1/2,1]},1_{[0,1/3]},1_{[1/3,2/3,]},1_{[2/3,1]},\dots$ which goes through all intervals of the form $[m/n,(m+1)/n]$ where $0<m<n$ are natural numbers.  %We may consider these functions as random variables on the sample space $\Omega=[0,1]$, with the uniform probability measure.
%That this sequence converges in probability to $0$ is straightforward to check---$\text{Pr}(1_{[0,1]}\neq0)=1$, but $\text{Pr}(1_{[0,1/2]}\neq0)=\text{Pr}(1_{[1/2,]}\neq0)

\end{document}